\documentclass{article} 
\usepackage{iclr2017_conference,times}
\usepackage{hyperref}
\usepackage{url}
\usepackage{xcolor,colortbl}
\usepackage{float}
\usepackage{graphicx}
\usepackage{amsmath}
\usepackage{subfig}

\title{Semantic Noise Modeling for \\ Better Representation Learning}

\author{Hyo-Eun Kim\thanks{ Corresponding author }~ and Sangheum Hwang \\
Lunit Inc. \\
Seoul, South Korea \\
\texttt{\{hekim, shwang\}@lunit.io} \\
\AND
\textbf{Kyunghyun Cho} \\
Courant Institute of Mathematical Sciences and Centre for Data Science \\
New York University \\
New York, NY 10012, USA \\
\texttt{kyunghyun.cho@nyu.edu}
}

\begin{document}

\maketitle

\begin{abstract}
Latent representation learned from multi-layered neural networks via hierarchical feature abstraction enables recent success of deep learning. Under the deep learning framework, generalization performance highly depends on the learned latent representation which is obtained from an appropriate training scenario with a task-specific objective on a designed network model. In this work, we propose a novel latent space modeling method to learn better latent representation. We designed a neural network model based on the assumption that good base representation can be attained by maximizing the total correlation between the input, latent, and output variables. From the base model, we introduce a \textit{semantic noise modeling} method which enables class-conditional perturbation on latent space to enhance the representational power of learned latent feature. During training, latent vector representation can be stochastically perturbed by a modeled class-conditional additive noise while maintaining its original semantic feature. It implicitly brings the effect of semantic augmentation on the latent space. The proposed model can be easily learned by back-propagation with common gradient-based optimization algorithms. Experimental results show that the proposed method helps to achieve performance benefits against various previous approaches. We also provide the empirical analyses for the proposed class-conditional perturbation process including t-SNE visualization.
\end{abstract}

\section{Introduction}
\label{introduction}
Enhancing the generalization performance against unseen data given some sample data is the main objective in machine learning. Under that point of view, deep learning has been achieved many breakthroughs in several domains such as computer vision~\citep{r01_alexnet_nips2012,r02_vggnet_iclr2015,r03_resnet_cvpr2016}, natural language processing~\citep{r04_nlp_icml2008,r05_nmt_iclr2015}, and speech recognition~\citep{r06_speech_magazine2012,r07_speech_icassp2013}. Deep learning is basically realized on deep layered neural network architecture, and it learns appropriate task-specific latent representation based on given training data. Better latent representation learned from training data results in better generalization over the future unseen data. Representation learning or latent space modeling becomes one of the key research topics in deep learning. During the past decade, researchers focused on unsupervised representation learning and achieved several remarkable landmarks on deep learning history~\citep{r08_sdae_jmlr2010,r09_dbn_neuralcomputation2006,r10_rbm_aistats2009}. In terms of utilizing good base features for supervised learning, the base representation learned from unsupervised learning can be a good solution for supervised tasks~\citep{r11_pretrain_nips2007,r12_pretrain_icann2011}. 

The definition of `good' representation is, however, different according to target tasks. In unsupervised learning, a model is learned from unlabelled examples. Its main objective is to build a model to estimate true data distribution given examples available for training, so the learned latent representation normally includes broadly-informative components of the raw input data (e.g., mutual information between the input and the latent variable can be maximized for this objective). In supervised learning, however, a model is learned from labelled examples. In the case of classification, a supervised model learns to discriminate input data in terms of the target task using corresponding labels. Latent representation is therefore obtained to maximize the performance on the target supervised tasks. 

Since the meaning of good representations vary according to target tasks (unsupervised or supervised), pre-trained features from the unsupervised model are not be guaranteed to be useful for subsequent supervised tasks. Instead of the two stage learning strategy (unsupervised pre-training followed by supervised fine-tuning), several works focused on a joint learning model which optimizes unsupervised and supervised objectives concurrently, resulting in better generalization performance~\citep{r13_jointlearning_nips2013,r14_jointlearning_icml2008,r15_jointlearning_nips2015,r16_jointlearning_iclr2016workshop,r17_jointlearning_icml2016,r22_jointlearning_cho2014}. 

In this work, we propose a novel latent space modeling method for supervised learning. We define a good latent representation of standard feed-forward neural networks under the basis of information theory. Then, we introduce a \textit{semantic noise modeling} method in order to enhance the generalization performance. The proposed method stochastically perturbs the latent representation of a training sample by injecting class-conditional additive noise. Since the additive noise is randomly sampled from a pre-defined probability distribution every training iteration, different latent vectors from a single training example can be used for training. The multiple different latent vectors produced from a single training example are semantically similar under the proposed class-conditional perturbation process, so we can expect semantic augmentation effect on the latent space. 

Experiments are performed on two datasets; MNIST and CIFAR-10. The proposed model results in better classification performance compared to previous approaches through notable generalization effect (class-conditionally perturbed training samples well cover the distribution of unseen data). 

\section{Methodology}
\label{methodology}

\begin{figure}[t]
\begin{center}
\includegraphics[width=0.75\textwidth]{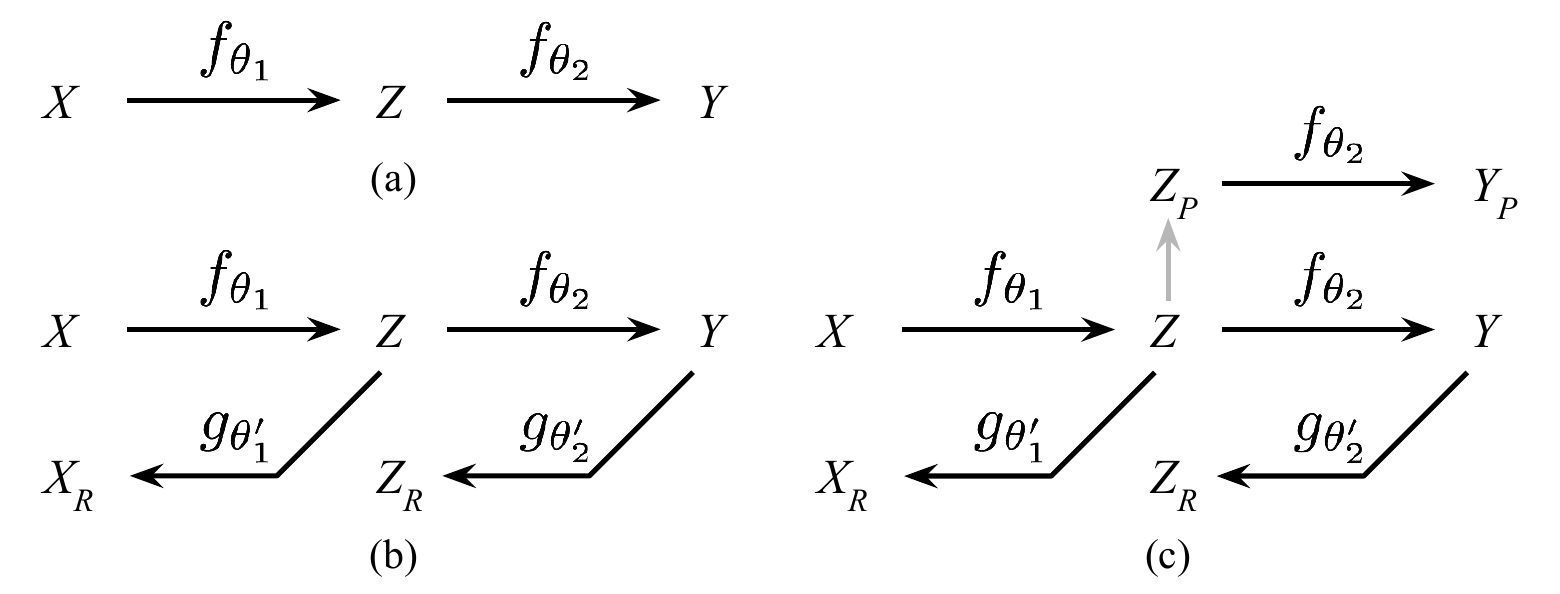}
\caption{(a) Standard feed-forward neural network model, (b) feed-forward neural network model with reconstruction paths, and (c) feed-forward neural network model with reconstruction and stochastic perturbation paths.}
\label{fig1:architectures}
\end{center}
\end{figure}

In a traditional feed-forward neural network model (Figure~\ref{fig1:architectures}(a)), output $Y$ of input data $X$ is compared with its true label, and the error is propagated backward from top to bottom, which implicitly learns a task-specific latent representation $Z$ of the input $X$. 
We assume that good latent representation $Z$ is attained by maximizing the dependency among a set of random variables $X$, $Y$, and $Z$, which is known as total correlation or multiinformation~\citep{r22_watanabe1960information}. Note that the total correlation is equal to the sum of all pairwise mutual informations. The total correlation $\textbf{C}(X,Y,Z)$ for given random variables  $X$, $Y$, and $Z$ under the condition $P(X,Y,Z) = P(Y \vert Z) P(Z \vert X) P(X)$ from the relationship between the random variables (in Figure~\ref{fig1:architectures}(a)) can be reduced to:

\begin{equation} \label{eq1}
\begin{split}
\textbf{C}(X,Y,Z)	& = \textbf{H}(X) + \textbf{H}(Y) + \textbf{H}(Z) - \textbf{H}(X,Y,Z) \\
						& = \textbf{H}(X) + \textbf{H}(Y) + \textbf{H}(Z) - (\textbf{H}(Y \vert Z) + \textbf{H}(Z \vert X) + \textbf{H}(X)) \\
                         	& = \textbf{I}(X;Z) + \textbf{I}(Z;Y) \\
                         	& = \textbf{H}(X) - \textbf{H}(X \vert Z) + \textbf{H}(Z) - \textbf{H}(Z \vert Y) \\
\end{split}
\end{equation}

where $\textbf{I}(A; B)$ is the mutual information between random variables $A$ and $B$, and $\textbf{H}(A)$ is the entropy of a random variable $A$. Our objective is to find the model parameter $\theta$ which maximizes $\textbf{C}(X,Y,Z)$. Since $\textbf{H}(X)$ and $\textbf{H}(Z)$ are non-negative, and $\textbf{H}(X)$ is constant in this case, the lower bound on $\textbf{C}(X,Y,Z)$ can be summarized as:

\begin{equation} \label{eq2}
\textbf{C}(X,Y,Z)	\geq - \textbf{H}(X \vert Z) - \textbf{H}(Z \vert Y)~.
\end{equation}

It is known that maximizing $-\textbf{H}(X \vert Z)$ can be formulated as minimizing the reconstruction error between the input $x$ (sampled from $X$) and its reconstruction $x_R$ under the general audo-encoder framework~\citep{r08_sdae_jmlr2010}. Similarly, maximizing $-\textbf{H}(Z \vert Y)$ can be reformulated by minimizing the reconstruction error between $z$ and its reconstruction $z_R$. The target objective can then be defined as follows:

\begin{equation} \label{eq3}
\min_{\theta} \lambda_1 L_{rec}(x, x_R) + \lambda_2 L_{rec}(z, z_R)
\end{equation}

where $\theta$ and $\lambda_{1,2}$ are model parameters to be learned and constant coefficients, and $L_{rec}$ is a reconstruction loss. 

Given an input sample $x$, feed-forwarded vectors and their reconstructions are attained deterministically by:

\begin{equation} \label{eq4}
\begin{split}
& z = f_{\theta_1}(x) \\
& y = f_{\theta_2}(f_{\theta_1}(x)) \\
& x_R = g_{\theta_1'}(z) = g_{\theta_1'}(f_{\theta_1}(x)) \\
& z_R = g_{\theta_2'}(y) = g_{\theta_2'}(f_{\theta_2}(f_{\theta_1}(x))
\end{split}
\end{equation}

where $x_R$ and $z_R$ are the reconstruction of $x$ and $z$ as shown in Figure~\ref{fig1:architectures}(b). 

For supervised learning, given a set of training pairs ($x$, $t$) where $x$ and $t$ are the input sample and its label, target objective under the model described in Figure~\ref{fig1:architectures}(b) can be defined as below (with real-valued input samples, L2 loss $L_{L2}$ is a proper choice for the reconstruction loss $L_{rec}$):

\begin{equation} \label{eq5}
\min_{\theta: \{\theta_1, \theta_1', \theta_2, \theta_2'\}}
 \lambda_1 L_{L2}(x, x_R) + \lambda_2 L_{L2}(z, z_R) + \lambda_3 L_{NLL}(y, t)
\end{equation}

where $L_{NLL}$ and $\lambda_3$ are a negative log-likelihood loss for the target supervised task and a relative weighting factor for $L_{NLL}$, respectively. Note that Eq.~(\ref{eq5}) represents the `\textit{proposed-base}' in our experiment (see Section~\ref{quantitative}).

Based on the architecture shown in Figure~\ref{fig1:architectures}(b) with the target objective in Eq.~(\ref{eq5}), we conjecture that stochastic perturbation on the latent space during training helps to achieve better generalization performance for supervised tasks. Figure~\ref{fig1:architectures}(c) shows this strategy which integrates the stochastic perturbation process during training. Suppose that $Z_P$ is a perturbed version of $Z$, and $Y_P$ is an output which is feed-forwarded from $Z_P$. Given an input sample $x$, 

\begin{equation} \label{eq6}
z' = z + z_e~~\text{and}~~
\hat{y} = f_{\theta_2}(z')
\end{equation} 

where $z'$ and $\hat{y}$ are samples of $Z_P$ and $Y_P$ respectively, and $z_e$ is an additive noise used in the perturbation process of $z$. Based on the architecture shown in Figure~\ref{fig1:architectures}(c), target objective can be modified as:

\begin{equation} \label{eq7}
\min_{\theta: \{\theta_1, \theta_1', \theta_2, \theta_2'\}}
 \lambda_1 L_{L2}(x, x_R) + \lambda_2 L_{L2}(z, z_R) + \lambda_3 L_{NLL}(y, t) + L_{NLL}(\hat{y}, t)~.
\end{equation}

Direct random additive noise is not appropriate for $z_e$ (`\textit{proposed-perturb (random)}' in Section~\ref{quantitative}), since random perturbation might destroy the semantic feature of the original latent representation $z$. In order to maintain the semantics of the original latent representation after perturbation, we design a class-conditional additive noise which can be modeled based on the architecture described in Figure~\ref{fig1:architectures}(b). We assume that the probability density function $P(Y_{(j)} \vert X)$ is approximately \textit{Gaussian} with the deterministic feed-forwarded value $y_{(j)}$ as a mean as below:

\begin{equation} \label{eq8}
P(Y_{(j)} \vert X) = \mathcal{N}(f_{\theta_2}(f_{\theta_1}(x))_{(j)},~\sigma_{(j)}^2) = y_{(j)} + \mathcal{N}(0, \sigma_{(j)}^2)
\end{equation}

where $Y_{(j)}$ and $\sigma_{(j)}$ are the $j$-th element of the random vector $Y$ and a standard deviation for $Y_{(j)}$. Now, the class-conditionally perturbed $z$ (i.e. $z'$ in Eq.~(\ref{eq6})) can be reconstructed from the class-conditionally perturbed $y$ (i.e. $y'$) through the decoding path $g_{\theta_2'}$. The semantic-preserving variation of $y$ (i.e. $y'$) can be modeled according to Eq.~(\ref{eq8}) by $y' = y + y_e$, where $y_e$ is a random noise vector which is stochastically sampled from the \textit{Gaussian} distribution. From $y'$, class-conditional additive noise on the latent space, $z_e$ (`\textit{proposed-perturb (class-conditional)}' in Section~\ref{quantitative}), can be approximately modeled as below:

\begin{equation} \label{eq9}
\begin{split}
& z_R = g_{\theta_2'}(y) \\
& z_R' = g_{\theta_2'}(y') = g_{\theta_2'}(y+y_e) \\
& z_e \simeq z_R' - z_R = g_{\theta_2'}(y+y_e) - g_{\theta_2'}(y)~.
\end{split}
\end{equation} 

From the described \textit{semantic noise modeling} process, we expect to achieve better representation on the latent space. The effect of the proposed model in terms of learned latent representation will be explained in more detail in Section~\ref{qualitative}.

\section{Related works}
\label{related}

\begin{figure}[t]
\begin{center}
\includegraphics[width=0.7\textwidth]{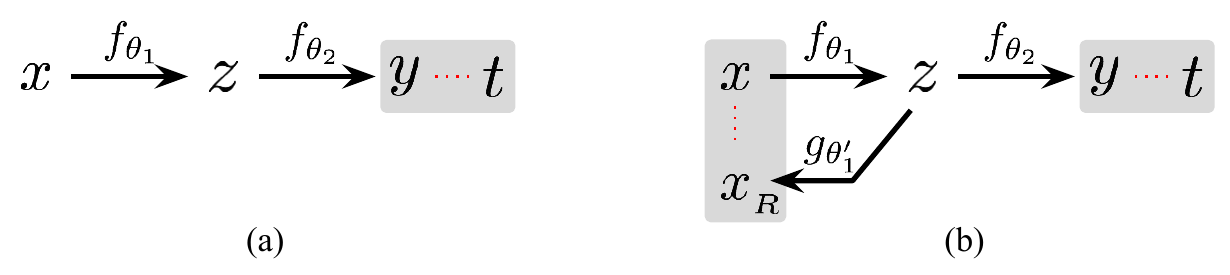}
\caption{Previous works for supervised learning; (a) traditional feed-forward model, and (b) joint learning model with both supervised and unsupervised losses.}
\label{fig2:related_arch}
\end{center}
\end{figure}

Previous works on deep neural networks for supervised learning can be categorized into two types as shown in Figure~\ref{fig2:related_arch}; (a) a general feed-forward neural network model~\citep{r18_lenet_ieeeproceeding1998,r01_alexnet_nips2012,r02_vggnet_iclr2015,r03_resnet_cvpr2016}, and (b) a joint learning model which optimizes unsupervised and supervised objectives at the same time~\citep{r16_jointlearning_iclr2016workshop,r17_jointlearning_icml2016,r22_jointlearning_cho2014}. Here are the corresponding objective functions:

\begin{equation} \label{eq10}
\min_{\theta: \{\theta_1, \theta_2\}} L_{NLL}(y, t)
\end{equation} 

\begin{equation} \label{eq11}
\min_{\theta: \{\theta_1, \theta_1', \theta_2\}} \lambda L_{L2}(x, x_R) + L_{NLL}(y, t)
\end{equation} 

where $\lambda$ is a loss weighting factor between unsupervised and supervised losses. 

Since the feed-forward neural network model is normally implemented with multiple layers in a deep learning framework, the joint learning model can be sub-classified into two types according to the type of reconstruction; reconstruction only with the input data $x$ (Eq.~(\ref{eq11})) and reconstruction with all the intermediate features including the input data $x$ as follows:

\begin{equation} \label{eq12}
\min_{\theta} \lambda_0 L_{L2}(x, x_R) + \sum_{i} \lambda_i L_{L2}(h_i, h_{i_R}) + L_{NLL}(y, t)~.
\end{equation} 

where $h_i$ and $h_{i_R}$ are the $i$-th hidden representation and its reconstruction.

\begin{figure}[t]
\begin{center}
\includegraphics[width=0.4\textwidth]{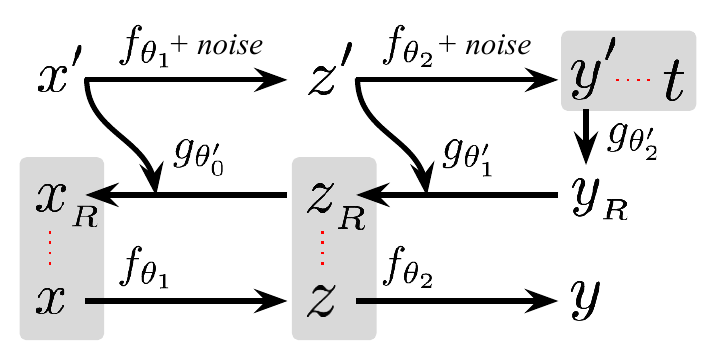}
\caption{Ladder network; a representative model for semi-supervised learning~\citep{r15_jointlearning_nips2015}.}
\label{fig3:ladder}
\end{center}
\end{figure}

Another type of the joint learning model, a \textit{ladder network} (Figure~\ref{fig3:ladder}), was introduced for semi-supervised learning~\citep{r15_jointlearning_nips2015}. The key concept of the ladder network is to obtain robust features by learning de-noising functions ($g_{\theta'}$) of the representations at every layer of the model via reconstruction losses, and the supervised loss is combined with the reconstruction losses in order to build the semi-supervised model. The ladder network achieved the best performance in semi-supervised tasks, but it is not appropriate for supervised tasks especially with small-scale training samples (experimental analysis for supervised learning on MNIST is briefly summarized in Appendix (A2)). The proposed model in this work can be extended to semi-supervised learning, but our main focus is to enhance the representational power on latent space given labelled data for supervised learning. We leave the study for semi-supervised learning scenario based on the proposed methodology as our future research.

\section{Experiments}
\label{experiments}
For quantitative analysis, we compare the proposed methodology with previous approaches described in Section~\ref{related}; a traditional feed-forward supervised learning model and a joint learning model with two different types of reconstruction losses (reconstruction only with the first layer or with all the intermediate layers including the first layer). The proposed methodology includes a baseline model in Figure~\ref{fig1:architectures}(b) as well as a stochastic perturbation model in Figure~\ref{fig1:architectures}(c). Especially in the stochastic perturbation model, we compare the random and class-conditional perturbations and present some qualitative analysis on the meaning of the proposed perturbation methodology.

\subsection{Datasets}
\label{dataset}
We experiment with two public datasets; MNIST and CIFAR-10. MNIST (10 classes) consists of 50k, 10k, and 10k 28$\times$28 gray-scale images for training, validation, and test datasets, respectively. CIFAR-10 (10 classes) consists of 50k and 10k 32$\times$32 3-channel images for training and test sets, respectively. We split the 50k CIFAR-10 training images into 40k and 10k for training and validation. Experiments are performed with different sizes of training set (from 10 examples per class to the entire training set) in order to verify the effectiveness of the proposed model in terms of generalization performance under varying sizes of training set.

\subsection{Implementation}
\label{implementation}
\begin{figure}[t]
\begin{center}
\includegraphics[width=0.7\textwidth]{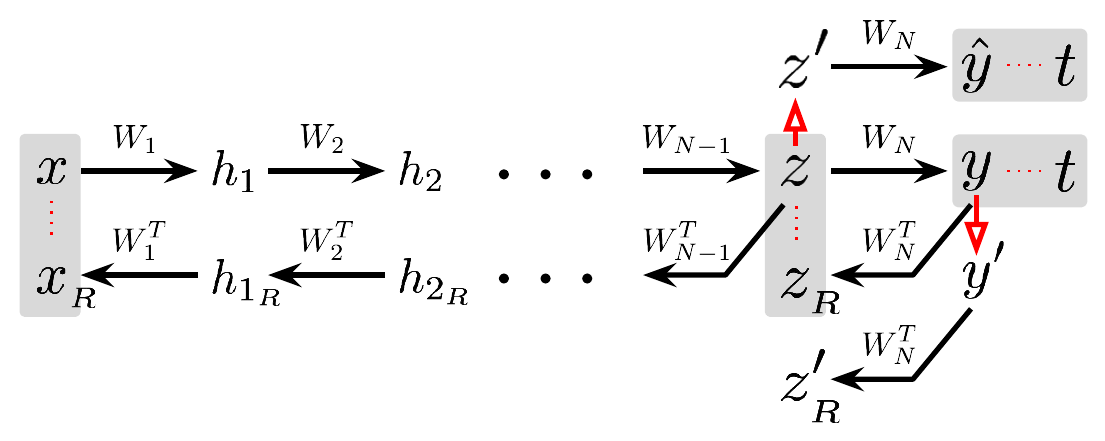}
\caption{Target network architecture; 3 convolution and 2 fully-connected layers were used for MNIST, and 4 convolution and 3 fully-connected layers were used for CIFAR-10.}
\label{fig4:specific_arch}
\end{center}
\end{figure}

Figure~\ref{fig4:specific_arch} shows the architecture of the neural network model used in this experiment. $W$'s are convolution or fully-connected weights (biases are excluded for visual brevity). Three convolution (3$\times$3 (2) 32, 3$\times$3 (2) 64, 3$\times$3 (2) 96, where each item means the filter kernel size and (stride) with the number of filters) and two fully-connected (the numbers of output nodes are 128 and 10, respectively) layers are used for MNIST. Four convolution (5$\times$5 (1) 64, 3$\times$3 (2) 64, 3$\times$3 (2) 64, and 3$\times$3 (2) 96) and three fully-connected (128, 128, and 10 nodes) layers are used for CIFAR-10. Weights on the decoding (reconstruction) path are tied with corresponding weights on the encoding path as shown in Figure~\ref{fig4:specific_arch}. 

In Figure~\ref{fig4:specific_arch}, $z'$ is perturbed directly from $z$ by adding \textit{Gaussian} random noise for random perturbation. For class-conditional perturbation, $z'$ is indirectly generated from $y'$ which is perturbed by adding random noise on $y$ based on Eq.~(\ref{eq9}). For perturbation, base activation vector ($z$ is the base vector for random perturbation and $y$ is the base vector for class-conditional perturbation) is scaled to [0.0, 1.0], and the zero-mean \textit{Gaussian} noise with 0.2 of standard deviation is added (via element-wise addition) on the normalized base activation. This perturbed scaled activation is de-scaled with the original min and max activations of the base vector.

Initial learning rates are 0.005 and 0.002 for MNIST and CIFAR-10, respectively. The learning rates are decayed by a factor of 5 every 40 epochs until the 120-th epoch. For both datasets, the minibatch size is set to 100, and the target objective is optimized using \textit{Adam} optimizer~\citep{r19_adam_iclr2015} with a momentum 0.9. All the $\lambda$'s for reconstruction losses in Eq.~(\ref{eq11}) and Eq.~(\ref{eq12}) are 0.03 and 0.01 for MNIST and CIFAR-10, respectively. The same weighting factors for reconstruction losses (0.03 for MNIST and 0.01 for CIFAR-10) are used for $\lambda_1$ and $\lambda_2$ in Eq~(\ref{eq7}), and 1.0 is used for $\lambda_3$. 

Input data is first scaled to [0.0, 1.0] and then whitened by the average across all the training examples. In CIFAR-10, random cropping (24$\times$24 image is randomly cropped from the original 32$\times$32 image) and random horizontal flipping (mirroring) are used for data augmentation. We selected the network that performed best on the validation dataset for evaluation on the test dataset. All the experiments are performed with TensorFlow~\citep{r21_tensorflow2015-whitepaper}.

\subsection{Quantitative analysis}
\label{quantitative}

Table~\ref{table1:performance} shows the classification performance of previous approaches and the proposed methods. Three previous approaches (a traditional feed-forward model, a joint learning model with the input reconstruction loss, and a joint learning model with reconstruction losses of all the intermediate layers including the input layer) are compared with three proposed methods (the baseline model in Figure~\ref{fig1:architectures}(b), and the stochastic perturbation model in Figure~\ref{fig1:architectures}(c) with two different perturbation methods; random and class-conditional).

As we expected, maximizing the total correlation (\textit{proposed-base}) learns better latent representation, and the model with the class-conditional perturbation (\textit{proposed-perturb (class-conditional)}) performs best among all the comparison targets. Especially in MNIST, the error rate of `\textit{proposed-perturb (class-conditional)}' with 2k per-class training examples is less than the error rate of all types of previous works with the entire training set (approximately 5k per-class examples). More empirical analyses in terms of the generalization performance are handled in next subsection.


\definecolor{Gray}{gray}{0.85}
\setlength{\tabcolsep}{6pt}
\begin{table}[t]
\caption{Error rate (\%) on the test set using the model with the best performance on the validation set. Numbers on the first row of each sub-table are the number of randomly chosen per-class training examples. The average performance of three different random-split datasets is described in this table (error rate on each random set is summarized in Appendix). Performance of three previous approaches (with gray background) and the proposed methods (baseline, random perturbation, and class-conditional perturbation in order) is summarized.}
\label{table1:performance}
\def\arraystretch{1.8}
\begin{center}\scriptsize
	\begin{tabular}{ l r r r r r r r r r }
	\hline\hline
	\textbf{MNIST} (\# train examples per class)											& 10 		& 20 		& 50 		& 100 	& 200 	& 500 	& 1k 		& 2k 		& (all) 5k 	\\
	\hline
	\rowcolor{Gray}
	feed-forward model; Figure~\ref{fig2:related_arch}(a)								& 24.55	& 16.00	& 10.35	& 6.58	& 4.71	& 2.94	& 1.90	& 1.45	& 1.04		\\
	\rowcolor{Gray}
	joint learning model with recon-one; Figure~\ref{fig2:related_arch}(b) 		& 21.67	& 13.60	& 7.85	& 5.44	& 4.14	& 2.50	& 1.84	& 1.45	& 1.12		\\
	\rowcolor{Gray}
	joint learning model with recon-all; Figure~\ref{fig2:related_arch}(b)		& \textbf{20.11}	& 13.69	& 9.15	& 6.77	& 5.39	& 3.89	& 2.91	& 2.28	& 1.87		\\
	proposed-base; Figure~\ref{fig1:architectures}(b)									& 21.35	& 11.65	& 6.33	& 4.32	& 3.07	& 1.98	& 1.29	& 0.94	& 0.80		\\
	proposed-perturb (random); Figure~\ref{fig1:architectures}(c)					& 20.17	& 11.68	& 6.24	& 4.12	& 3.04	& 1.88	& 1.24	& 0.96	& 0.65		\\
	proposed-perturb (class-conditional); Figure~\ref{fig1:architectures}(c)	& \textbf{20.11}	& \textbf{10.59}	& \textbf{5.92}	&\textbf{ 3.79}	& \textbf{2.72}	& \textbf{1.78}	& \textbf{1.15}	& \textbf{0.88}	& \textbf{0.62}		\\
	\hline\hline
	\textbf{CIFAR-10} (\# train examples per class)										& 10 		& 20 		& 50 		& 100 	& 200 	& 500 	& 1k 		& 2k 		& (all) 4k 	\\
	\hline
	\rowcolor{Gray}
	feed-forward model; Figure~\ref{fig2:related_arch}(a)								& 73.82	& 68.99	& 61.30	& 54.93	& 46.97	& 33.69	& 26.63	& 20.97	&	17.80	\\
	\rowcolor{Gray}
	joint learning model with recon-one; Figure~\ref{fig2:related_arch}(b) 		& 75.68	& 69.05	& 61.44	& 55.02	& 46.18	& 33.62	& 26.78	& 21.25	&	17.68	\\
	\rowcolor{Gray}
	joint learning model with recon-all; Figure~\ref{fig2:related_arch}(b)		& 73.33	& 67.63	& 62.59	& 56.37	& 50.51	& 41.26	& 32.55	& 26.38	&	22.71	\\
	proposed-base; Figure~\ref{fig1:architectures}(b)									& 71.63	& \textbf{66.17}	& 58.91	& 52.65	& 43.46	& 31.86	& 25.76	& 21.06	&	17.45	\\
	proposed-perturb (random); Figure~\ref{fig1:architectures}(c)					& 71.69	& 66.75	& 58.95	& 53.01	& 43.71	& 31.80	& 25.50	& 20.81	&	17.43	\\
	proposed-perturb (class-conditional); Figure~\ref{fig1:architectures}(c)	& \textbf{71.50}	& 66.87	& \textbf{58.30}	& \textbf{52.32}	& \textbf{42.98}	& \textbf{30.91}	& \textbf{24.81}	& \textbf{20.19}	&	\textbf{16.16}	\\
	\hline\hline
	\end{tabular}
\end{center}
\end{table}
\setlength{\tabcolsep}{1.4pt}

\subsection{Qualitative analysis}
\label{qualitative}

\begin{figure}[t]
\begin{center}
\includegraphics[width=1.0\textwidth]{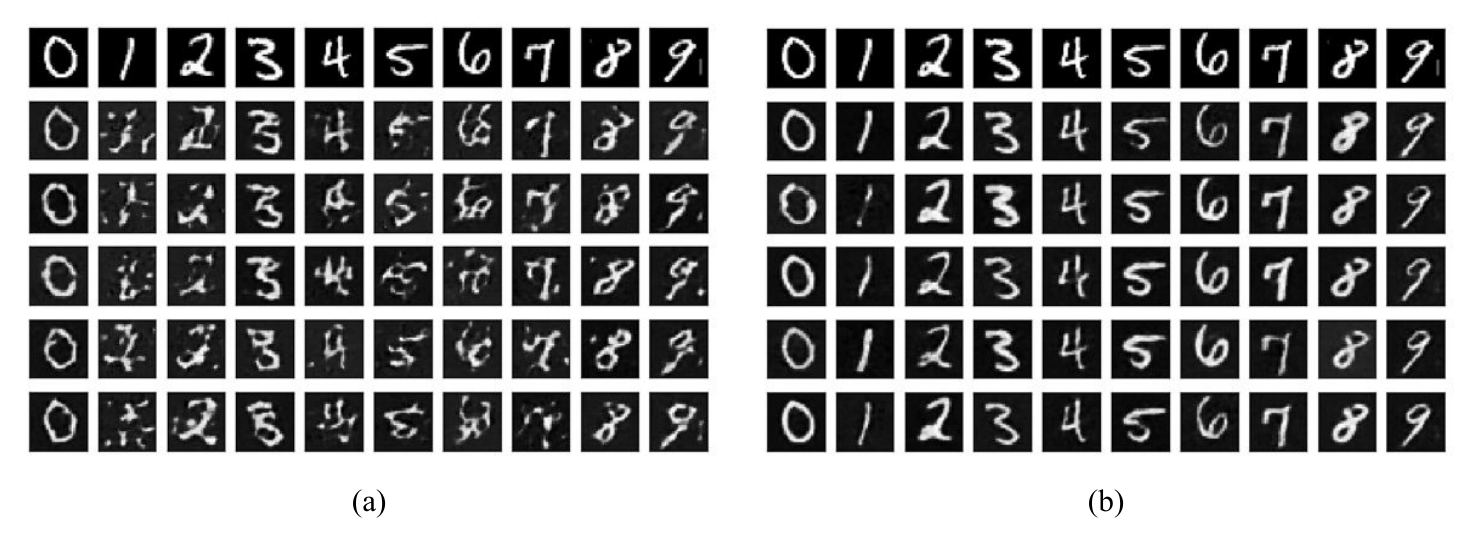}
\caption{Examples reconstructed from (a) randomly, and (b) class-conditionally perturbed latent vectors (top row shows the original training examples).}
\label{fig5:samples}
\end{center}
\end{figure}

As mentioned before, random perturbation by adding unstructured noise directly to the latent representation easily destroys the semantic feature of the original representation. We compared two different perturbation methods (random and class-conditional) by visualizing the examples reconstructed from the perturbed latent vectors (Figure~\ref{fig5:samples}). Top row is the original examples selected from training set (among 2k per-class training examples), and the rest are the reconstructions of their perturbed latent representations. Based on the architecture described in Figure~\ref{fig1:architectures}(b), we generated five different perturbed latent representations according to the type of perturbation, and reconstructed the perturbed latent vectors through decoding path for reconstruction.

Figure~\ref{fig5:samples}(a) and (b) show the examples reconstructed from the random and class-conditional perturbations, respectively. For both cases, zero-mean \textit{Gaussian} random noise (0.2 standard deviation) is used for perturbation. As shown in Figure~\ref{fig5:samples}(a), random perturbation cannot guarantee the preservation of original semantics; for example, semantics of `1' is mostly destroyed under random perturbation, and some examples of `3' are reconstructed as being similar to `8' rather than its original content `3'. Figure~\ref{fig5:samples}(b) shows the examples reconstructed from the class-conditionally perturbed representation. The reconstructed examples show subtle semantic variations while maintaining the original semantic contents; for example, thickness difference in `3' (example on the third row) or writing style difference in `8' (openness of the top left corner).

\begin{figure}[t]
\begin{center}
\includegraphics[width=1.0\textwidth]{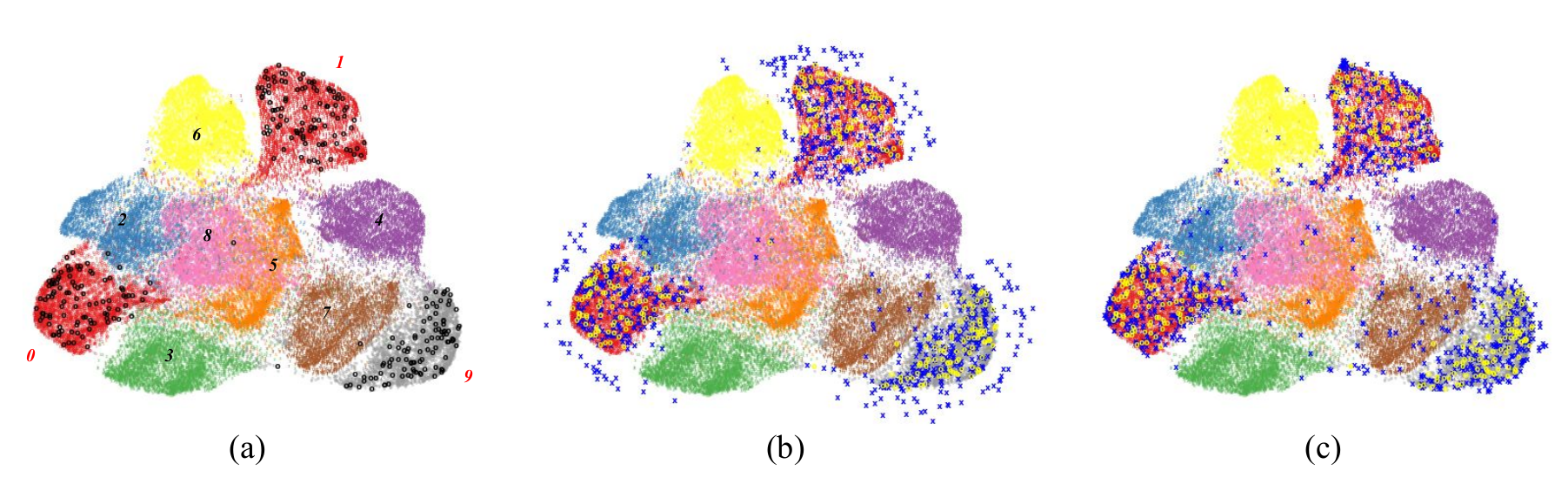}
\caption{Training examples (circles or crosses with colors described below) over the examples not used for training (depicted as background with different colors); (a) training examples (black circles), (b) training examples (yellow circles) with 3$\times$ randomly perturbed samples (blue crosses), and (c) training examples (yellow circles) with 3$\times$ class-conditionally perturbed samples (blue crosses). Best viewed in color.}
\label{fig6:tsne}
\end{center}
\end{figure}

Figure~\ref{fig6:tsne} shows the overall effect of the perturbation. In this analysis, 100 per-class MNIST examples are used for training. From the trained model based on the architecture described in Figure~\ref{fig1:architectures}(b), latent representations $z$ of all the 50k examples (among 50k examples, only 1k examples were used for training) are visualized by using t-SNE~\citep{r20_tsne_jmlr2008}. Only the training examples of three classes (0, 1, and 9) among ten classes are depicted as black circles for visual discrimination in Figure~\ref{fig6:tsne}(a). The rest of the examples which were not used for training (approximately 4.9k examples per class) are depicted as a background with different colors. We treat the colored background examples (not used for training) as a true distribution of unseen data in order to estimate the generalization level of learned representation according to the type of perturbation. Figure~\ref{fig6:tsne}(b) and (c) show the training examples (100 examples per class with yellow circles) and their perturbed ones (3$\times$ sampled from each example with blue crosses) through random and class-conditional perturbations, respectively. 

In Figure~\ref{fig6:tsne}(b), perturbed samples are distributed near the original training examples, but some samples outside the true distribution cannot be identified easily with appropriate classes. This can be explained with Figure~\ref{fig5:samples}(a), since some perturbed samples are ambiguous semantically. In Figure~\ref{fig6:tsne}(c), however, most of the perturbed samples evenly cover the true distribution. As mentioned before, stochastic perturbation with the class-conditional additive noise during training implicitly incurs the effect of augmentation on the latent space while resulting in better generalization. Per-class t-SNE results are summarized in Appendix (A3.1).

\section{Discussion}
\label{discussion}
We introduced a novel latent space modeling method for supervised tasks based on the standard feed-forward neural network architecture. The presented model simultaneously optimizes both supervised and unsupervised losses based on the assumption that the better latent representation can be obtained by maximizing the total correlation of all the random variables defined by the standard feed-forward neural network model. Especially the stochastic perturbation process which is achieved by modeling the class-conditional additive noise during training enhances the representational power of the latent space. From the proposed \textit{semantic noise modeling} process, we can expect improvement of generalization performance in supervised learning with implicit semantic augmentation effect on the latent space.

The presented model architecture can be intuitively extended to semi-supervised learning because it is implemented as the joint optimization of supervised and unsupervised objectives. For semi-supervised learning, however, logical link between features learned from labelled and unlabelled data needs to be considered additionally. We leave the extension of the presented approach to semi-supervised learning for the future.



\bibliography{iclr2017_conference}
\bibliographystyle{iclr2017_conference}

\newpage
\section*{Appendix}
\label{appendix}

\subsection*{(A1) Quantitative analysis}
\textbf{Extended from Section~\ref{quantitative}.} 
Among the total 50k and 40k training examples in MNIST and CIFAR-10, we randomly select the examples for training. Classification performance according to three different randomly chosen training sets are summarized in Table~\ref{table_a1:MNIST} (MNIST) and Table~\ref{table_a2:CIFAR-10} (CIFAR-10). Further experiments with denoising constraints are also included. Zero-mean \textit{Gaussian} random noise with 0.1 standard deviation is used for noise injection. Denoising function helps to achieve slightly better performance on MNIST, but it results in performance degradation on CIFAR-10 (we did not focus on searching the optimal parameters for noise injection in this experiments).

\definecolor{Gray}{gray}{0.85}
\setlength{\tabcolsep}{5pt}
\begin{table}[H]
\caption{Classification performance (error rate in \%) on three different sets of randomly chosen training examples (MNIST).}
\label{table_a1:MNIST}
\def\arraystretch{1.8}
\begin{center}\scriptsize
	\begin{tabular}{ l r r r r r r r r r }
	\hline\hline
	\textbf{Set No.1} (\# train examples per class)											& 10 		& 20 		& 50 		& 100 	& 200 	& 500 	& 1k 		& 2k 		& (all) 5k 	\\
	\hline
	\rowcolor{Gray}
	feed-forward model; Figure~\ref{fig2:related_arch}(a)								& 22.61	& 14.20	& 11.25	& 6.37	& 4.34	& 2.63	& 1.83	& 1.56	& 1.04		\\
	\rowcolor{Gray}
	joint learning model with recon-one; Figure~\ref{fig2:related_arch}(b) 		& 18.69	& 12.21	& 7.84	& 5.17	& 4.02	& 2.58	& 1.79	& 1.47	& 1.12		\\
	\rowcolor{Gray}
	joint learning model with recon-one with denoising constraints 				& 20.39	& 11.91	& 7.41	& 4.64	& 3.65	& 2.57	& 1.97	& 1.53	& 0.97		\\
	\rowcolor{Gray}
	joint learning model with recon-all; Figure~\ref{fig2:related_arch}(b)		& 18.82	& 12.82	& 9.34	& 6.43	& 5.23	& 4.12	& 2.68	& 2.42	& 1.87		\\
	\rowcolor{Gray}
	joint learning model with recon-all with denoising constraints 					& 17.93	& 11.76	& 7.32	& 4.78	& 3.91	& 3.04	& 2.52	& 1.99	& 1.36		\\
	proposed-base; Figure~\ref{fig1:architectures}(b)									& 20.23	& 10.18	& 6.47	& 3.89	& 3.04	& 1.89	& 1.33	& 0.91	& 0.80		\\
	proposed-base with denoising constraints 											& 19.88	& 10.89	& 6.62	& 4.26	& 3.40	& 2.44	& 2.11	& 1.54	& 1.13		\\
	proposed-perturb (random); Figure~\ref{fig1:architectures}(c)					& 18.38	& 10.58	& 6.64	& 3.78	& 3.14	& 1.90	& 1.21	& 0.89	& 0.65		\\
	proposed-perturb (class-conditional); Figure~\ref{fig1:architectures}(c)	& 19.33	& 9.72	& 5.98	& 3.47	& 2.84	& 1.84	& 1.16	& 0.84	& 0.62		\\
	\hline\hline
	\textbf{Set No.2} (\# train examples per class)											& 10 		& 20 		& 50 		& 100 	& 200 	& 500 	& 1k 		& 2k 		&  	\\
	\hline
	\rowcolor{Gray}
	feed-forward model; Figure~\ref{fig2:related_arch}(a)								& 28.84	& 17.36	& 10.14	& 6.20	& 4.78	& 3.02	& 1.61	& 1.41	& 		\\
	\rowcolor{Gray}
	joint learning model with recon-one; Figure~\ref{fig2:related_arch}(b)	 	& 26.09	& 14.40	& 7.98	& 5.18	& 4.17	& 2.29	& 1.94	& 1.52	& 		\\
	\rowcolor{Gray}
	joint learning model with recon-one with denoising constraints 				& 27.69	& 13.11	& 6.95	& 5.07	& 3.54	& 2.37	& 1.83	& 1.28	& 		\\
	\rowcolor{Gray}
	joint learning model with recon-all; Figure~\ref{fig2:related_arch}(b)		& 24.01	& 14.13	& 8.98	& 6.84	& 5.44	& 3.51	& 2.98	& 2.18	& 		\\
	\rowcolor{Gray}
	joint learning model with recon-all with denoising constraints 					& 23.05	& 13.29	& 7.79	& 5.12	& 3.92	& 3.01	& 2.27	& 1.84	& 		\\
	proposed-base; Figure~\ref{fig1:architectures}(b)									& 22.95	& 12.98	& 6.27	& 4.43	& 3.22	& 2.14	& 1.37	& 0.96	& 		\\
	proposed-base with denoising constraints 											& 26.96	& 12.21	& 6.45	& 4.62	& 3.13	& 2.53	& 1.88	& 1.49	& 		\\
	proposed-perturb (random); Figure~\ref{fig1:architectures}(c)					& 22.10	& 12.52	& 5.97	& 4.26	& 2.86	& 1.94	& 1.23	& 0.92	& 		\\
	proposed-perturb (class-conditional); Figure~\ref{fig1:architectures}(c)	& 21.22	& 11.52	& 5.75	& 3.91	& 2.61	& 1.73	& 1.14	& 0.89	& 		\\
	\hline\hline
	\textbf{Set No.3} (\# train examples per class)											& 10 		& 20 		& 50 		& 100 	& 200 	& 500 	& 1k 		& 2k 		&  	\\
	\hline
	\rowcolor{Gray}
	feed-forward model; Figure~\ref{fig2:related_arch}(a)								& 22.20	& 16.43	& 9.67	& 7.16	& 5.02	& 3.17	& 2.25	& 1.39	& 		\\
	\rowcolor{Gray}
	joint learning model with recon-one; Figure~\ref{fig2:related_arch}(b)	 	& 20.23	& 14.19	& 7.73	& 5.96	& 4.22	& 2.62	& 1.79	& 1.35	& 		\\
	\rowcolor{Gray}
	joint learning model with recon-one with denoising constraints 				& 19.32	& 12.25	& 7.44	& 5.39	& 3.58	& 2.37	& 1.49	& 1.56	& 		\\
	\rowcolor{Gray}
	joint learning model with recon-all; Figure~\ref{fig2:related_arch}(b)		& 17.51	& 14.12	& 9.12	& 7.04	& 5.49	& 4.05	& 3.08	& 2.25	& 		\\
	\rowcolor{Gray}
	joint learning model with recon-all with denoising constraints 					& 17.07	& 12.50	& 7.86	& 5.48	& 4.05	& 2.97	& 2.02	& 1.98	& 		\\
	proposed-base; Figure~\ref{fig1:architectures}(b)									& 20.86	& 11.79	& 6.25	& 4.63	& 2.96	& 1.91	& 1.16	& 0.96	& 		\\
	proposed-base with denoising constraints 											& 19.89	& 11.30	& 6.26	& 4.57	& 3.50	& 2.63	& 1.61	& 1.47	& 		\\
	proposed-perturb (random); Figure~\ref{fig1:architectures}(c)					& 20.02	& 11.94	& 6.12	& 4.32	& 3.13	& 1.81	& 1.28	& 1.08	& 		\\
	proposed-perturb (class-conditional); Figure~\ref{fig1:architectures}(c)	& 19.78	& 10.53	& 6.03	& 4.00	& 2.70	& 1.76	& 1.14	& 0.92	& 		\\
	\hline\hline
	\end{tabular}
\end{center}
\end{table}
\setlength{\tabcolsep}{1.4pt}

\definecolor{Gray}{gray}{0.85}
\setlength{\tabcolsep}{5pt}
\begin{table}[H]
\caption{Classification performance (error rate in \%) on three different sets of randomly chosen training examples (CIFAR-10).}
\label{table_a2:CIFAR-10}
\def\arraystretch{1.8}
\begin{center}\scriptsize
	\begin{tabular}{ l r r r r r r r r r }
	\hline\hline
	\textbf{Set No.1} (\# train examples per class)											& 10 		& 20 		& 50 		& 100 	& 200 	& 500 	& 1k 		& 2k 		& (all) 4k 	\\
	\hline
	\rowcolor{Gray}
	feed-forward model; Figure~\ref{fig2:related_arch}(a)								& 73.30	& 69.25	& 62.42	& 55.65	& 47.71	& 34.30	& 27.04	& 21.06	& 17.80		\\
	\rowcolor{Gray}
	joint learning model with recon-one; Figure~\ref{fig2:related_arch}(b)	 	& 75.19	& 70.38	& 62.25	& 55.30	& 46.89	& 34.12	& 26.63	& 21.05	& 17.68		\\
	\rowcolor{Gray}
	joint learning model with recon-one with denoising constraints 				& 73.72	& 68.20	& 61.99	& 55.23	& 46.64	& 36.37	& 29.78	& 25.53	& 21.73		\\
	\rowcolor{Gray}
	joint learning model with recon-all; Figure~\ref{fig2:related_arch}(b)		& 74.79	& 68.33	& 62.92	& 56.24	& 51.37	& 40.30	& 30.91	& 26.49	& 22.71		\\
	\rowcolor{Gray}
	joint learning model with recon-all with denoising constraints 					& 76.56	& 69.67	& 64.53	& 57.88	& 52.74	& 42.24	& 36.90	& 30.93	& 27.41		\\
	proposed-base; Figure~\ref{fig1:architectures}(b)									& 70.79	& 66.57	& 59.91	& 52.98	& 43.29	& 32.25	& 26.19	& 20.92	& 17.45		\\
	proposed-base with denoising constraints 											& 71.03	& 67.49	& 60.37	& 53.52	& 44.28	& 33.40	& 28.00	& 25.06	& 21.34		\\
	proposed-perturb (random); Figure~\ref{fig1:architectures}(c)					& 71.89	& 67.12	& 59.22	& 52.79	& 43.87	& 31.82	& 25.04	& 20.97	& 17.43		\\
	proposed-perturb (class-conditional); Figure~\ref{fig1:architectures}(c)	& 71.59	& 66.90	& 58.64	& 52.34	& 42.74	& 30.94	& 24.45	& 20.10	& 16.16		\\
	\hline\hline
	\textbf{Set No.2} (\# train examples per class)											& 10 		& 20 		& 50 		& 100 	& 200 	& 500 	& 1k 		& 2k 		&  	\\
	\hline
	\rowcolor{Gray}
	feed-forward model; Figure~\ref{fig2:related_arch}(a)								& 72.39	& 69.49	& 60.45	& 54.85	& 46.91	& 33.39	& 26.73	& 21.00	& 		\\
	\rowcolor{Gray}
	joint learning model with recon-one; Figure~\ref{fig2:related_arch}(b)	 	& 74.06	& 69.14	& 60.71	& 54.54	& 45.70	& 33.54	& 27.43	& 20.90	& 		\\
	\rowcolor{Gray}
	joint learning model with recon-one with denoising constraints 				& 76.40	& 69.33	& 60.28	& 55.38	& 47.40	& 36.29	& 29.31	& 24.60	& 		\\
	\rowcolor{Gray}
	joint learning model with recon-all; Figure~\ref{fig2:related_arch}(b)		& 72.28	& 67.60	& 61.53	& 56.65	& 49.99	& 42.08	& 32.99	& 26.33	& 		\\
	\rowcolor{Gray}
	joint learning model with recon-all with denoising constraints 					& 73.90	& 69.23	& 61.90	& 57.99	& 52.35	& 45.12	& 37.23	& 30.14	& 		\\
	proposed-base; Figure~\ref{fig1:architectures}(b)									& 72.49	& 65.62	& 57.82	& 52.66	& 43.20	& 32.24	& 25.60	& 21.32	& 		\\
	proposed-base with denoising constraints 											& 72.99	& 66.75	& 57.78	& 53.81	& 44.33	& 33.56	& 28.40	& 25.03	& 		\\
	proposed-perturb (random); Figure~\ref{fig1:architectures}(c)					& 71.84	& 65.98	& 58.08	& 53.37	& 43.44	& 31.56	& 25.69	& 21.03	& 		\\
	proposed-perturb (class-conditional); Figure~\ref{fig1:architectures}(c)	& 72.85	& 66.65	& 57.44	& 52.21	& 42.74	& 31.17	& 24.99	& 20.54	& 		\\
	\hline\hline
	\textbf{Set No.3} (\# train examples per class)											& 10 		& 20 		& 50 		& 100 	& 200 	& 500 	& 1k 		& 2k 		&  	\\
	\hline
	\rowcolor{Gray}
	feed-forward model; Figure~\ref{fig2:related_arch}(a)								& 75.78	& 68.24	& 61.02	& 54.29	& 46.28	& 33.38	& 26.11	& 20.85	& 		\\
	\rowcolor{Gray}
	joint learning model with recon-one; Figure~\ref{fig2:related_arch}(b)	 	& 77.79	& 67.62	& 61.37	& 55.22	& 45.96	& 33.21	& 26.29	& 21.81	& 		\\
	\rowcolor{Gray}
	joint learning model with recon-one with denoising constraints 				& 76.60	& 69.27	& 61.13	& 55.10	& 47.50	& 37.12	& 29.63	& 24.88	& 		\\
	\rowcolor{Gray}
	joint learning model with recon-all; Figure~\ref{fig2:related_arch}(b)		& 72.92	& 66.97	& 63.31	& 56.23	& 50.16	& 41.41	& 33.75	& 26.31	& 		\\
	\rowcolor{Gray}
	joint learning model with recon-all with denoising constraints 					& 76.83	& 68.53	& 65.58	& 58.29	& 52.43	& 45.42	& 39.01	& 32.32	& 		\\
	proposed-base; Figure~\ref{fig1:architectures}(b)									& 71.60	& 66.31	& 58.99	& 52.30	& 43.88	& 31.10	& 25.48	& 20.95	& 		\\
	proposed-base with denoising constraints 											& 72.39	& 67.20	& 60.60	& 52.64	& 44.62	& 33.52	& 28.01	& 25.25	& 		\\
	proposed-perturb (random); Figure~\ref{fig1:architectures}(c)					& 71.34	& 67.15	& 59.55	& 52.86	& 43.81	& 32.01	& 25.78	& 20.42	& 		\\
	proposed-perturb (class-conditional); Figure~\ref{fig1:architectures}(c)	& 70.06	& 67.07	& 58.83	& 52.41	& 43.47	& 30.61	& 25.00	& 19.94	& 		\\
	\hline\hline
	\end{tabular}
\end{center}
\end{table}
\setlength{\tabcolsep}{1.4pt}

\subsection*{(A2) Ladder network, a representative semi-supervised learning model}
\textbf{Extended from Section~\ref{related}.} 
We performed experiments with a ladder network model~\citep{r15_jointlearning_nips2015} in order to estimate the performance on supervised tasks according to different sizes of training set. We used the code (https://github.com/rinuboney/ladder.git) implemented with tensorflow~\citep{r21_tensorflow2015-whitepaper} for the experiment; the network architecture implemented on the source code is used as is; (784-1000-500-250-250-250-10). Although the base network architectures are different between ours and the ladder network, we can approximately gauge the relative performance difference between two models in supervised tasks (Table~\ref{table_a3:ladder}). Although the ladder network model works well for semi-supervised learning, it is not appropriate for fully supervised tasks especially with small-scale datasets.

\definecolor{Gray}{gray}{0.85}
\setlength{\tabcolsep}{6pt}
\begin{table}[H]
\caption{Classification performance (error rate in \%) of the ladder network and the proposed model on three different sets of randomly chosen training examples (MNIST).}
\label{table_a3:ladder}
\def\arraystretch{1.8}
\begin{center}\scriptsize
	\begin{tabular}{ l r r r r r r r r r }
	\hline\hline
	\textbf{Set No.1} (\# train examples per class)											& 10 		& 20 		& 50 		& 100 	& 200 	& 500 	& 1k 		& 2k 		& (all) 5k 	\\
	\hline
	\rowcolor{Gray}
	ladder network model; Figure~\ref{fig3:ladder}										& 25.85	& 16.48	& 9.26	& 6.00	& 4.66	& 3.07	& 2.15	& 1.26	& 0.91		\\
	proposed-perturb (class-conditional); Figure~\ref{fig1:architectures}(c)	& 19.33	& 9.72	& 5.98	& 3.47	& 2.84	& 1.84	& 1.16	& 0.84	& 0.62		\\
	\hline\hline
	\textbf{Set No.2} (\# train examples per class)											& 10 		& 20 		& 50 		& 100 	& 200 	& 500 	& 1k 		& 2k 		&  	\\
	\hline
	\rowcolor{Gray}
	ladder network model; Figure~\ref{fig3:ladder}										& 33.14	& 17.46	& 10.44	& 6.67	& 4.43	& 2.82	& 1.94	& 1.37	& 		\\
	proposed-perturb (class-conditional); Figure~\ref{fig1:architectures}(c)	& 21.22	& 11.52	& 5.75	& 3.91	& 2.61	& 1.73	& 1.14	& 0.89	& 		\\
	\hline\hline
	\textbf{Set No.3} (\# train examples per class)											& 10 		& 20 		& 50 		& 100 	& 200 	& 500 	& 1k 		& 2k 		&  	\\
	\hline
	\rowcolor{Gray}
	ladder network model; Figure~\ref{fig3:ladder}										& 29.99	& 16.99	& 9.73	& 7.34	& 4.39	& 3.00	& 2.12	& 1.47	& 		\\
	proposed-perturb (class-conditional); Figure~\ref{fig1:architectures}(c)	& 19.78	& 10.53	& 6.03	& 4.00	& 2.70	& 1.76	& 1.14	& 0.92	& 		\\
	\hline\hline
	\end{tabular}
\end{center}
\end{table}
\setlength{\tabcolsep}{1.4pt}

\subsection*{(A3.1) Qualitative analysis}
\textbf{Extended from Section~\ref{qualitative}.} 
Figure~\ref{fig7:tsne} shows the t-SNE results per class on MNIST. The overall tendency is similar to the description in Section~\ref{qualitative}.

\begin{figure}[H]
\begin{center}
\vskip -0.2in
  \begin{minipage}{\textwidth}
  \includegraphics[width=1.0\textwidth]{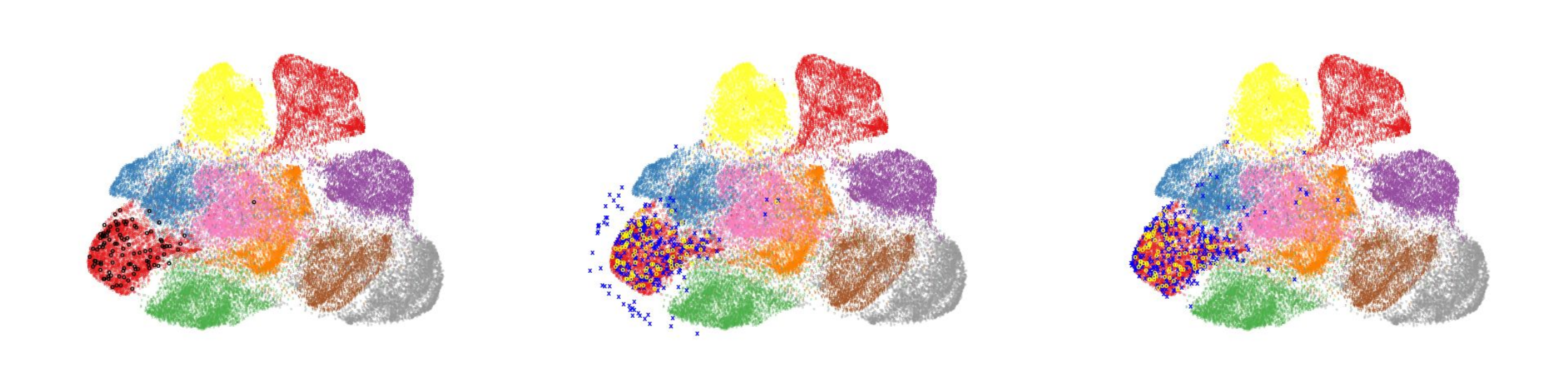}
  \end{minipage}
\vskip -0.2in
  \begin{minipage}{\textwidth}
  \includegraphics[width=1.0\textwidth]{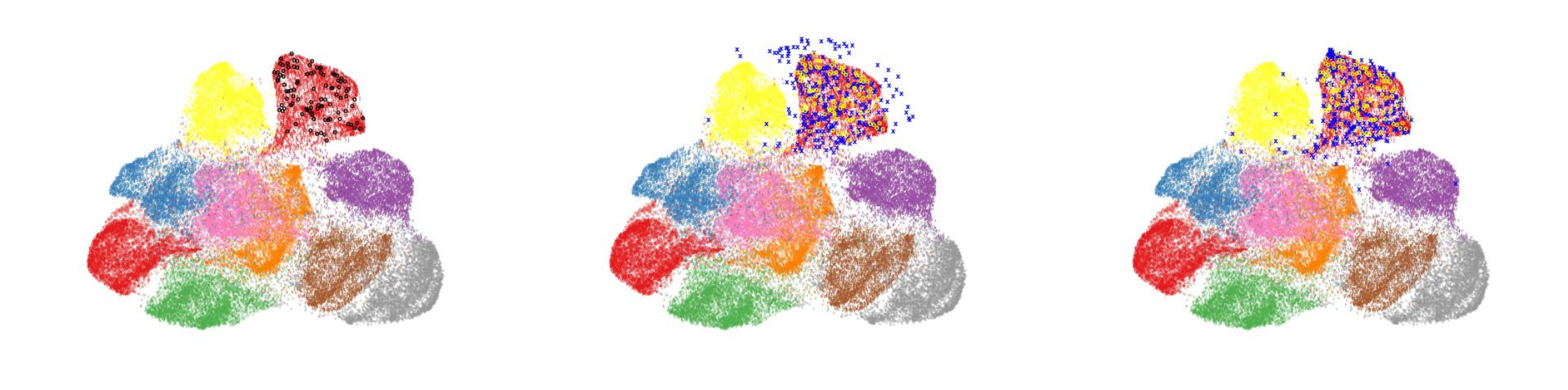}
  \end{minipage}
\vskip -0.2in
  \begin{minipage}{\textwidth}
  \includegraphics[width=1.0\textwidth]{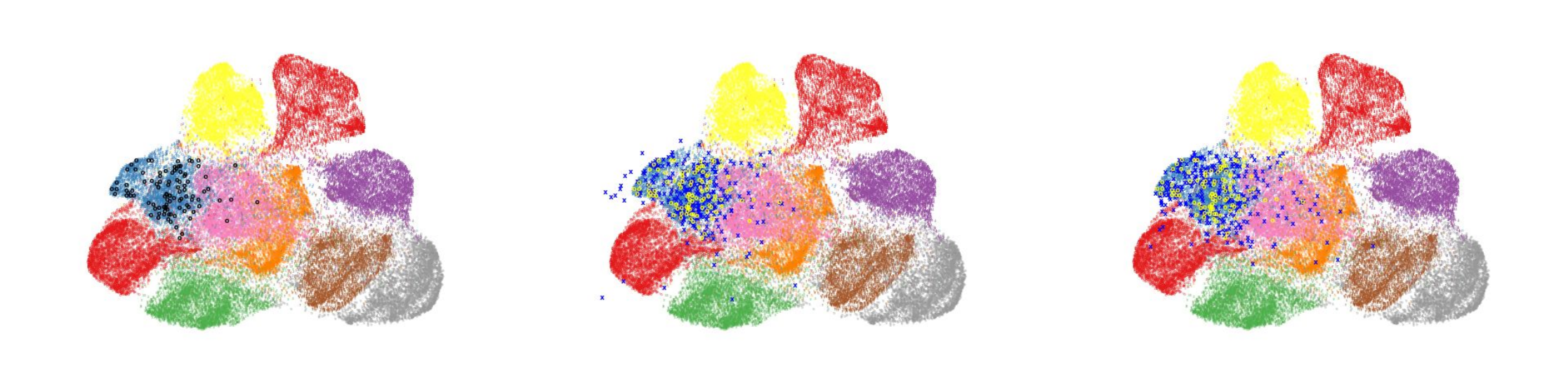}
  \end{minipage}
\vskip -0.2in
  \begin{minipage}{\textwidth}
  \includegraphics[width=1.0\textwidth]{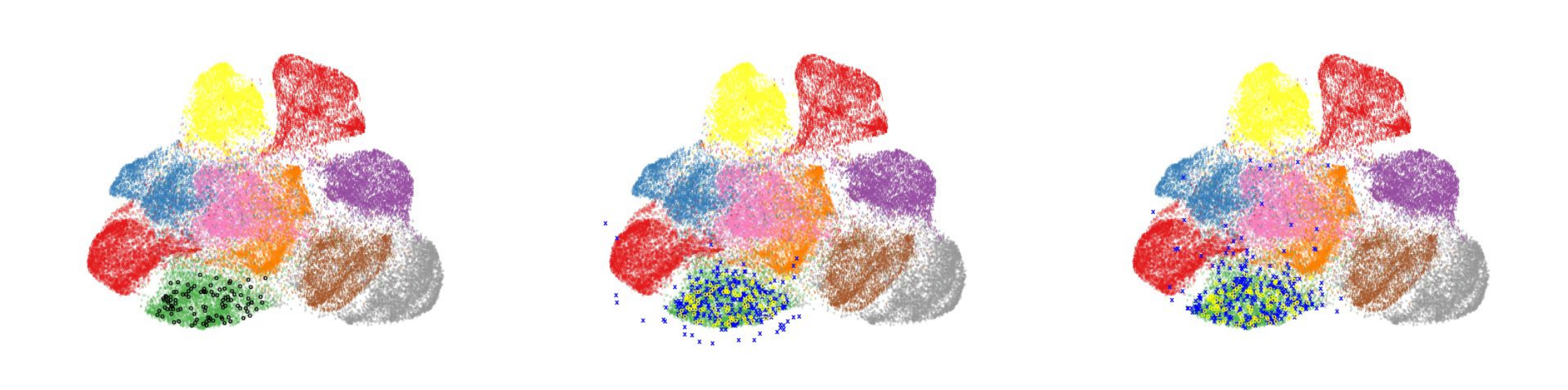}
  \end{minipage}
\vskip -0.2in
\end{center}
\end{figure}

\begin{figure}[H]
\begin{center}
\vskip -0.2in
  \begin{minipage}{\textwidth}
  \includegraphics[width=1.0\textwidth]{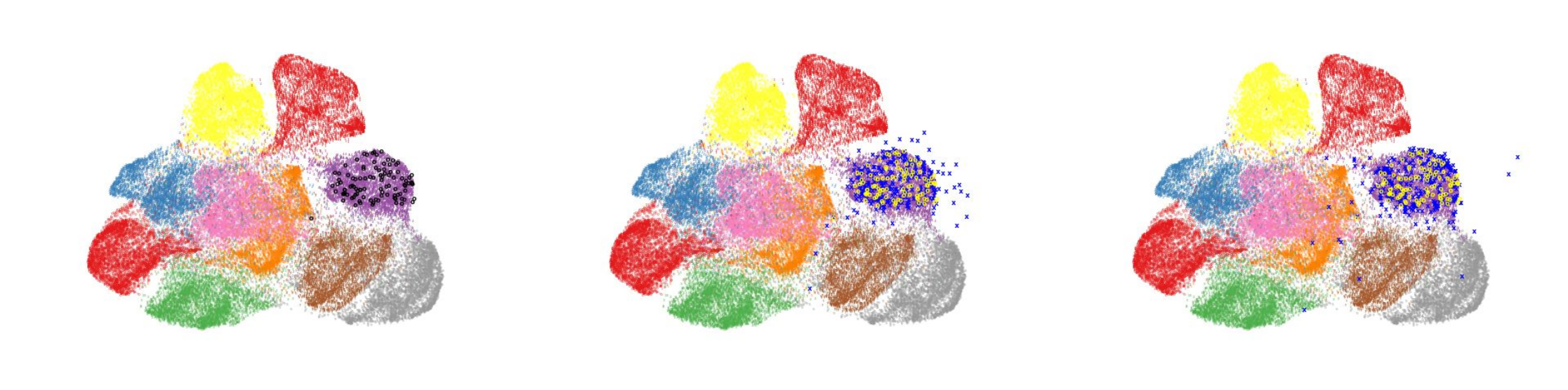}
  \end{minipage}
\vskip -0.2in
  \begin{minipage}{\textwidth}
  \includegraphics[width=1.0\textwidth]{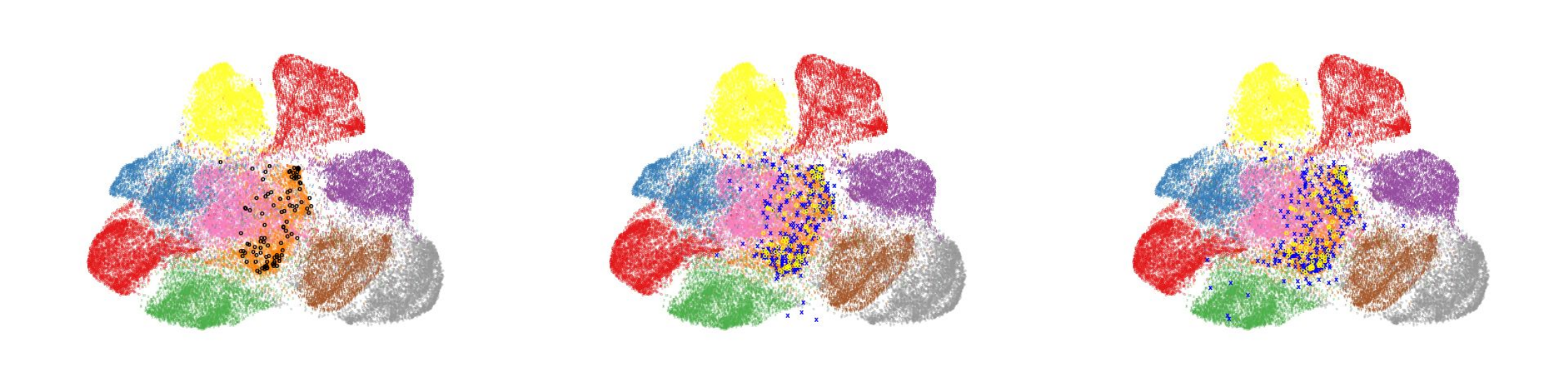}
  \end{minipage}
\vskip -0.2in
  \begin{minipage}{\textwidth}
  \includegraphics[width=1.0\textwidth]{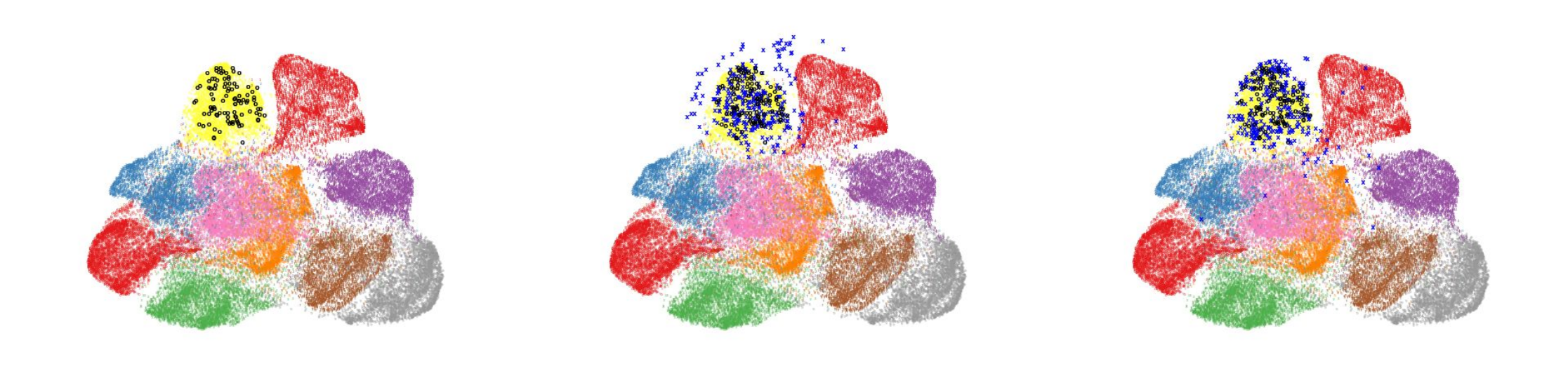}
  \end{minipage}
\vskip -0.2in
  \begin{minipage}{\textwidth}
  \includegraphics[width=1.0\textwidth]{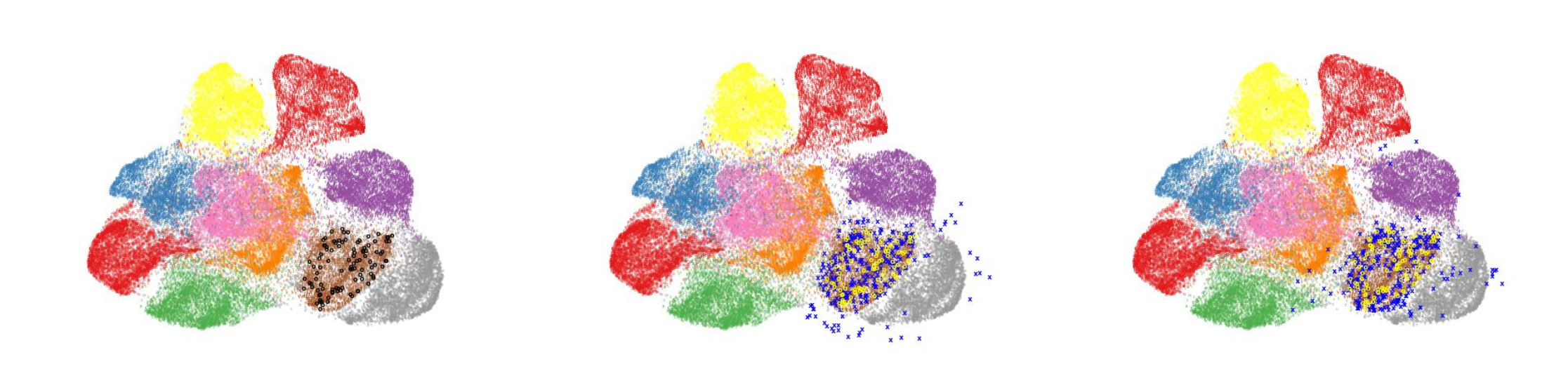}
  \end{minipage}
\vskip -0.2in
  \begin{minipage}{\textwidth}
  \includegraphics[width=1.0\textwidth]{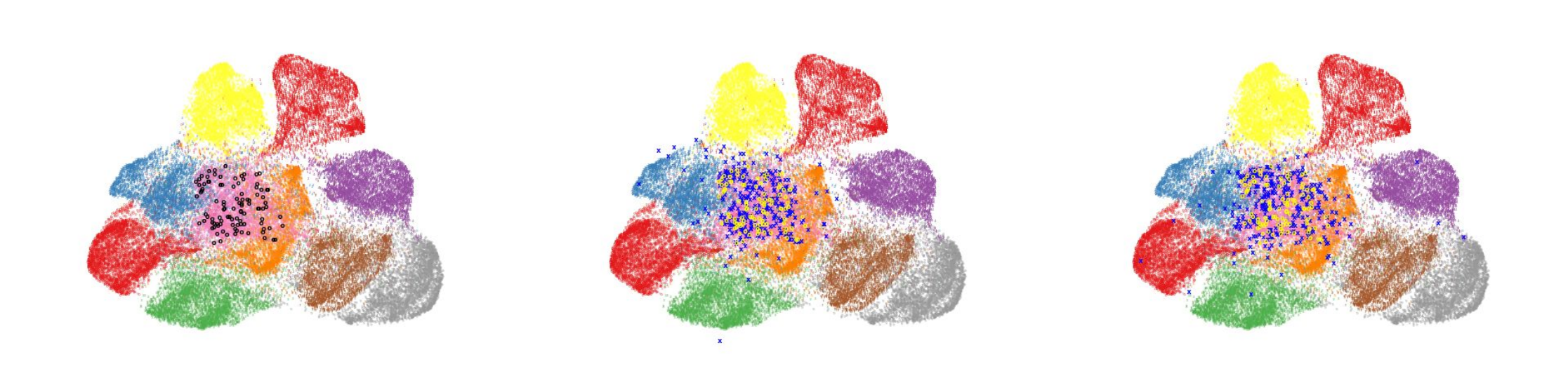}
  \end{minipage}
\vskip -0.2in
  \begin{minipage}{\textwidth}
  \includegraphics[width=1.0\textwidth]{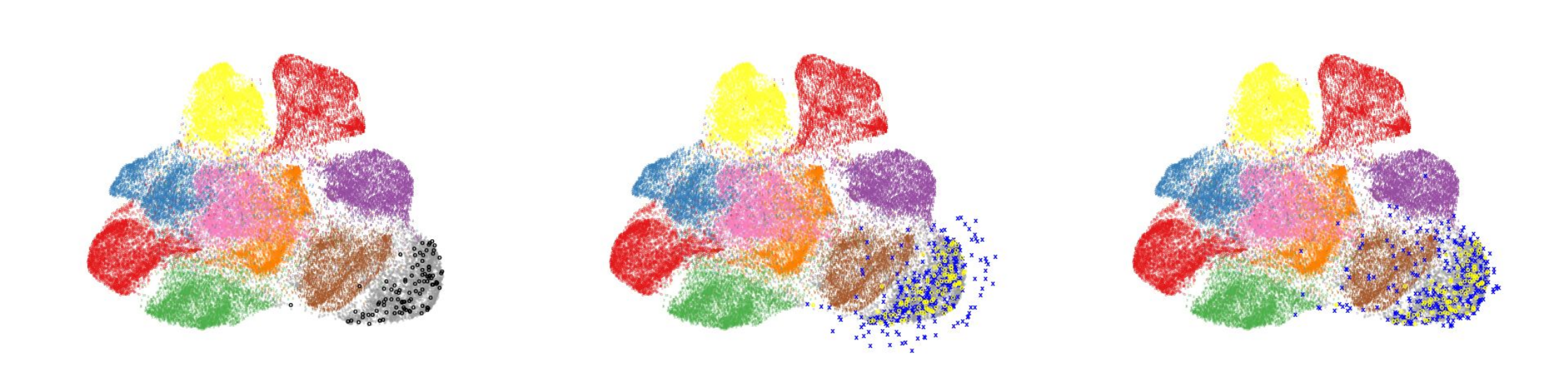}
  \end{minipage}
\caption{From top to bottom: 0, 1, 2, 3, 4, 5, 6, 7, 8, and 9. From left to right: training examples (circle), training examples (circle) + randomly perturbed samples (cross), and training examples (circle) + class-conditionally perturbed samples (cross). Best viewed in color.}
\label{fig7:tsne}
\end{center}
\end{figure}

\subsection*{(A3.2) Qualitative analysis}
\begin{figure}[H]
\begin{center}
  \begin{minipage}{\textwidth}
  \includegraphics[width=1.0\textwidth]{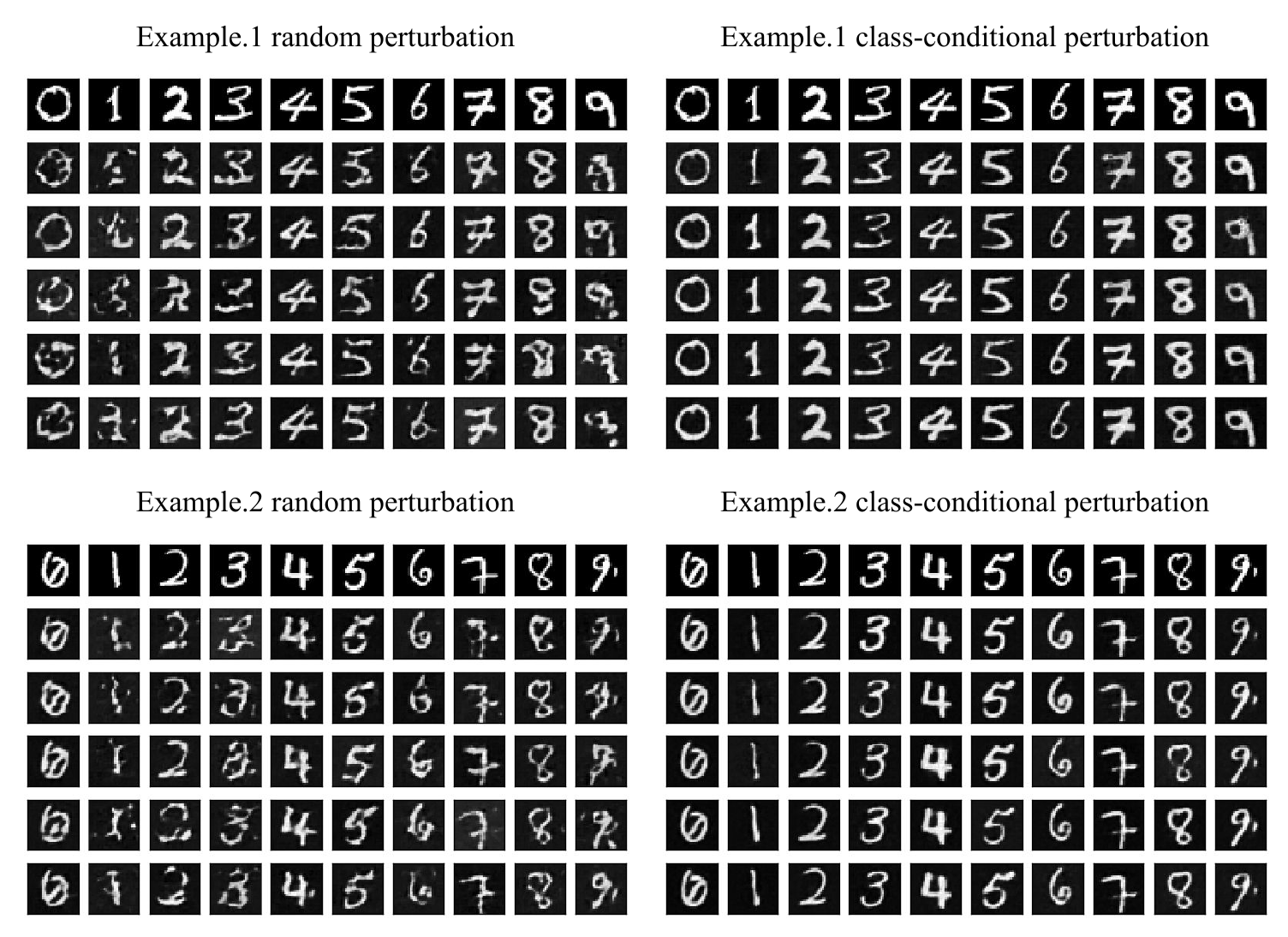}
  \end{minipage}
\caption{For each example, top row is the original examples selected from the training set, and the rest are reconstructed from the perturbed representations via random (left) and class-conditional (right) perturbations.}
\label{fig8:samples}
\end{center}
\end{figure}

\textbf{Extended from Section~\ref{qualitative}.} 
Figure~\ref{fig8:samples} shows reconstructed examples from perturbed (randomly or class-conditionally) latent representations (refer to Figure~\ref{fig5:samples} and the analysis described in Section~\ref{qualitative}).

\end{document}